\documentclass[review]{elsarticle}

\usepackage{lineno,hyperref}
\usepackage[printwatermark]{xwatermark}
\usepackage{xcolor}
\usepackage{graphicx}

\begin{document}

\begin{frontmatter}

\title{Entropy, Computing and Rationality}

\author{Luis A. Pineda\fnref{myfootnote}}
\address{Universidad Nacional Aut\'onoma de M\'exico}
\fntext[myfootnote]{lpineda@unam.mx}




\begin{abstract}

Making decisions freely presupposes that there is some indeterminacy in the environment and in the decision making engine. The former is reflected on the behavioral changes due to communicating: few changes indicate rigid environments; productive changes manifest a moderate indeterminacy, but a large communicating effort with few productive changes characterize a chaotic environment. Hence, communicating, effective decision making and productive behavioral changes are related. The entropy measures the indeterminacy of the environment, and there is an entropy range in which communicating supports effective decision making. This conjecture is referred to here as the \emph{The Potential Productivity of Decisions}. 

The computing engine that is causal to decision making should also have some indeterminacy. However, computations performed by standard Turing Machines are predetermined. To overcome this limitation an entropic mode of computing that is called here \emph{Relational-Indeterminate} is presented. Its implementation in a table format has been used to model an associative memory. The present theory and experiment suggest the \emph{Entropy Trade-off}: There is an entropy range in which computing is effective but if the entropy is too low computations are too rigid and if it is too high computations are unfeasible. The entropy trade-off of computing engines corresponds to the potential productivity of decisions of the environment.

The theory is referred to an Interaction-Oriented Cognitive Architecture. Memory, perception, action and thought involve a level of indeterminacy and decision making may be free in such degree. The overall theory supports an ecological view of rationality. The entropy of the brain has been measured in neuroscience studies and the present theory supports that the brain is an entropic machine. The paper is concluded with a number of predictions that may be tested empirically.

\end{abstract}


\end{frontmatter}

\newtheorem{mydef}{Definition}
\linenumbers

\nolinenumbers

\section{Potential Productivity of Decisions}
\label{sec:entropy}

People and machines make decisions all the time but the question is whether these are made freely or are rather predetermined. This is the old opposition between determinism and free will. The latter presupposes that there must be some indeterminacy either in the environment in which the decision is made or in the mind of the decision maker, or in both. 

The indeterminacy corresponds to the information content of the environment or \emph{the control volume}. Information is measured in communication theory with Shannon's entropy \cite{Shannon1948}:  $s = - \sum_{i=1}^{n}p(x_i) \times log_2(p(x_i))$. This is a formula of expected value. The term $log_2(p(x_i))$ is the length in bits of a message with probability $p(x_i)$; therefore, the entropy is the average length of a message in the communication environment. The probability of a message corresponds to the probability of the reported event. Events that are certain to occur have probability of one, and need not be communicated, or are communicated with ``messages of length zero''. The length of the message increases with its information content and the larger the entropy the greater the amount of information in the control volume. The information entropy reflects the uncertainty of the events that need to be communicated and, consequently, the indeterminacy of the environment.

Communication is achieved through signals carrying messages but while the former are physical phenomena the interpretation of the latter belongs to the plane of content. The entropy times the total number of messages in the control volume is proportional to the energy invested in communicating in such environment over a period of time; hence, the entropy reflects the overall effort that the community invest in communicating. These considerations are explicit or follow directly from Shannon's original presentation. Here, this discussion is extended to the relation between communication, decision making and behavioral change.

Communication allows individuals of a society --the family, the school, the office, the working institution, the church, the linguistic community, etc.-- to profit from the knowledge and experience of others. Communicative actions underlie the intention to change the knowledge, beliefs, desires, feelings, emotions, intentions and, most fundamentally, the course of action that the interlocutors would undergo without the information provided. Communicating presupposes that it is possible to change behavior. If the interlocutors are to deviate from their normal course of action on the basis of novel information they must have the choice; hence communication is a precondition for decision making in social environments.

Effective decision making reflects the indeterminacy of the physical but also the social environment. To enact a decision it must be physically and behaviorally possible: if the world or the society are too rigid, behavior cannot be changed, decisions cannot be enacted, communicating is not reinforced and the entropy is low. Conversely, productive behavioral changes due to communication reflect effective decision making and that the environment is less determined, with the corresponding larger entropy. However, large communication effort and high entropy but with few productive behavioral changes reflect that decision making is not effective and that the environment is too unpredictable or chaotic. These considerations suggest that the entropy is not only a measure of the indeterminacy of the physical environment but also of the plane of interpretation or content, and that there is a range of entropy in which decision making is effective. This conjecture is called here \emph{The Potential Productivity of Decisions}. 

The relation between the entropy of communication environments and the potential productivity of decisions, whose value is designated as $\tau$, is illustrated through the following scenarios, each defining a particular control volume:

\begin{itemize}
\item Factory production line: workers in a factory line do not communicate and if they do, communication does not affect the dynamics of the environment. The entropy and $\tau$ are very low or zero. There is no decision making.
\item Normal daily life: people communicate commonly to change beliefs and behavior of others; speech acts in normal conversation aim to achieve such an effect. The entropy is low or moderate and $\tau$ has an acceptable value. There is space for decision making.
\item Creative environments: people communicate effectively to make decisions that if enacted have important consequences. The entropy and $\tau$ are optimal. There is effective decision making with potential significant impact.
\item Crisis situation: an earthquake or a pandemic. People communicate a great deal and the entropy is very high but the environment is chaotic  and $\tau$ has a very low value. There may be a large decision making activity, but decisions are not enacted or do not achieve the intended effect or do it poorly. 
\end{itemize}

\begin{figure}
\includegraphics[width=0.5\textwidth]{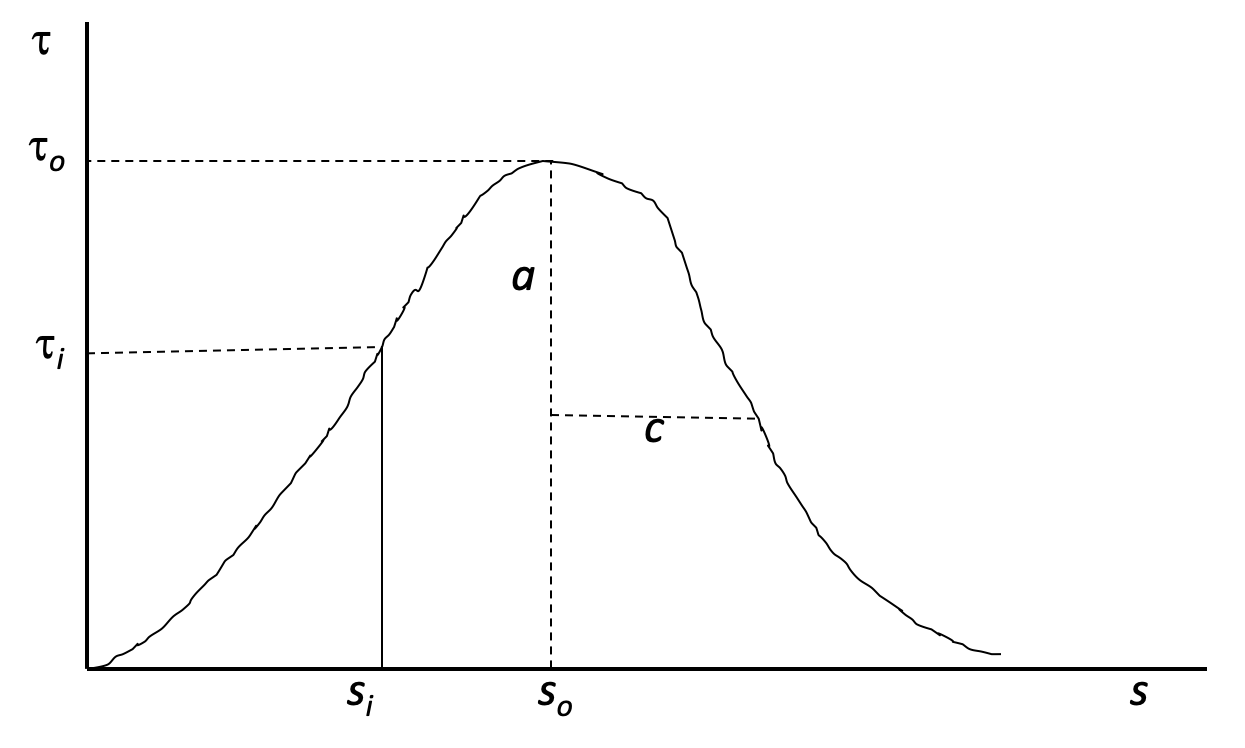}
\centering
\caption{Potential Productivity of Decisions}
\label{fig:productivity}
\end{figure}

The simplest characterization of the $\tau$ profile of environments is a gaussian function $\psi$ from the entropy $s$ to the potential productivity of decisions, such that $\tau = \psi(s) = ae^{-(s-s_0)^2/2c^2}$ where $s \geq 0$, $a$ is the value of $\tau$ at the optimal entropy $s_o$, and $c$ is the standard deviation. Each of the four scenarios above correspond to a particular control volume with entropy $s_i$ and productivity of decisions $\tau_i$. These are illustrated by shifting $s_i$ from left to right in Figure \ref{fig:productivity}.

The potential productivity of decisions can be seen as a cost-benefit parameter of communicating in a control volume. This is an ecological variable indicating the degree to which communicating provides an adaptation advantage and this behavior is reinforced. In the limiting case, when $\tau=0$, communication is of little use and agents may be better off by themselves, and when $\tau$ is too large communication is reduced to social noice. This conjecture may be tested empirically.

Communication and decision making can be performed by a wide variety of animal species that produce sounds or motor actions with communicative intent. Such ``speech acts'' can be codified through messages, although most species have a pretty determined ``view of the world'', have limited abilities to cope with environmental changes, and \emph{their} entropy and $\tau$ may be both very low. From an ecological and evolutive perspectives, the potential productivity of decisions measures the degree in which the plane of content matters for the species. Those that do not communicate or communicate little are limited to behave reactively to the signals sensed in the environment, and there is no reason to suppose that they sustain a plane of content or have a mind, and a large communication effort that is not accompanied by productive behavioral changes is not reinforced.

\section{Relational-Indeterminate Computing}
\label{sec:relational-computing}

The computing engine that is causal and essential to decision making must also have a choice. If such engine is deterministic, decisions are predetermined and decision making is an illusion. According to standard thinking in Computer Science the Turing Machine (TM) is the most general computing machine --all other general enough computing machines are equivalent to it-- and this machine is deterministic. In \emph{Computing Machinery and Intelligence} \cite{turing-1950} Turing stated that the predictions made by digital computers are ``rather nearer to practicality than that considered by Laplace'' (Turing, 1950, s.5). Laplace's demon is indeed a TM that computes all future and previous states of the universe given the full set of the physical laws and the initial conditions of a particular state, and it does so instantly. Hence, decisions ``made'' by digital computers are predetermined.

Computing machines hold physical states and interchange signals but do not make interpretations and decisions by themselves. Communication, information, knowledge and decision making are objects at plane of interpretation that is held in the mind of humans and of other animals with a developed enough neural system. For this, computing machinery is designed and used in relation to a standard configuration, which defines the format of representations, and a set of interpretation conventions.

The most basic interpretation convention in the theory of computability is that every TM computes a particular function, and the enumeration of all TMs correspond to the set of computational functions (e.g., \cite{boolos-jeffrey-1989}). A function is a relation between two arbitrary sets of objects, the domain and the codomain, whose members are called here \emph{the arguments} and \emph{the values} respectively, such that an argument is related to one value at the most, and this relation is given when the function is defined.

Mathematical objects are immutable, and the relation between the arguments and values of all functions is fixed. Algorithms are mechanical procedures that produce the value of the given argument, but the output of the computation is predetermined. Algorithms can be seen as intensional definitions of functions but the same knowledge can be expressed extensionally, as in tables, and what an algorithm does is simply to render explicitly the implicit knowledge. For all of this, under the standard interpretation conventions, TM computations are predetermined by necessity. Hence, the entropy of a TM is zero, and the notion of entropy is alien to the standard theory of TMs and computability.

The absolute determinism of TMs can be questioned from the perspective of the so-called non-deterministic automata \cite{hopcroft-2006}. There are non-deterministic finite state automata that have an equivalent deterministic one, such as those accepting regular languages; in this case, the non-determinism can be seen as a means to express abstractions but the actual computations are deterministic; however, there are automata that are genuinely non-deterministic, such as those accepting ambiguous languages or machines that explore a problem space heuristically, such as computer chess programs. In these latter cases the input or argument is associated to more than one value and the object that is computed is not a function but a mathematical relation. 

The determinism of TMs is also questioned from the perspective of stochastic computations. Turing suggested this strategy in the 1950 paper: when the problem space is too large, it can be partitioned in regions, and the value of the function for the given argument can be found by jumping into promising regions, using random numbers. His example was the problem of finding whether a number between 50 and 200 is equal to the sum of the squares of its constituent digits (Turing, 1950, s.7). This problem has no solution but illustrates the contrast between the deterministic strategy --iterating from the first to the last number in the domain-- and the stochastic one --choosing a number in the domain randomly and making the test until the solution is found. This latter process may never end, unless all trials are recorded and the computation finishes when all the instances of the problem have been tested, but with a very high cost in terms of memory and processing time. Turing called these methods the \emph{systematic} and the \emph{learning} respectively, and stated that the latter may be regarded as a search for a form of behavior and that ``since  there is probably a very large number of satisfactory solutions the random method seems to be better than the systematic'' (Turing, 1950, s.7). As there are many solutions, the object of computing is again a relation rather than a function.

The stochastic method requires that there is some indeterminacy in the computing process. This was also made explicit by Turing who proposed a variant of a digital computer with a random element, and that random numbers can be simulated in deterministic computers; for instance, by choosing the next digit in the expansion of the number $\pi$ (Turing, 1950, s.4). These digits are not known in advance but the sequence is predetermined, and such numbers are pseudo-random rather than random; hence, computations involving them are in the end determined. Genuine random numbers can be and are produced by sensing a property of the external environment --that cannot be predicted-- whose value can be seen as a hidden argument of the stochastic algorithm. Although such argument is not known, the object being computed is nevertheless a function, and the process is still deterministic.

This limitation can be overcome by generalizing the interpretation conventions and postulate that the object of computing is the mathematical relation rather than the function. A relation assigns an arbitrary number of objects of the codomain to each object in the domain. Hence, the value of the relation for a given argument is indeterminate. To address this indeterminacy the basic notion of \emph{evaluating a relation} is construed here as choosing randomly one among all the values of the relation for the given argument. This is: $R(a_i) = v_j$ such that $v_j$ is selected randomly --with an appropriate distribution-- among the values associated to the argument $a_i$ in the relation $R$. This latter interpretation convention is more general, computations become indeterminate, the computing engine is stochastic and the machine has an intrinsic computing entropy. This mode is called here \emph{Relational-Indeterminate Computing} (RIC).

The entropy of a relational-indeterminate computation is defined here as the normalized average indeterminacy of all the arguments of the relation $R$.
Let $\mu_i$ be the number of values associated to the argument $a_i$ in $R$, $\nu_i$ = $1/\mu_i$ 
and $n$ is the cardinality of the domain. In case $R$ is partial and $\mu_i = 0$ for the argument $a_i$ then 
$\nu_i = 1$. The computational entropy $e$ --or the entropy of $R$-- is defined here as $e(R) = -1/n \sum_{i=1}^{n} log_2(\nu_i)$.

The communication and computational entropies have a common normalized scale. A function has one value for all of its arguments and its entropy is zero. Partial functions do not define a value for all the arguments, but this is fully determined and the entropy of partial functions is also zero.

The relational-indeterminate mode of computing focuses on relational information and has three basic operations: \emph{abstraction}, \emph{containment} and \emph{reduction}. Let the sets $A = \{a_1,...,a_n\}$ and $V = \{v_1,...,n_m\}$, of cardinalities $n$ and $m$, be the domain and the codomain of a finite relation $R: A\to V$. For purposes of notation, for any relation $R$ a function with the same name is defined as follows: $R: A\times V\to \{0,1\}$ such that $R(a_i,v_j) = 1$ or \emph{true} if the argument $a_i$ is related to the value $v_j$ in $R$, and $R(a_i,v_j) = 0$ or \emph{false} otherwise.

RCI has three basic operations: \emph{abstraction}, \emph{containment} and \emph{reduction}. Let $R_f$ and $R_a$ be two arbitrary relations from $A$ to $V$, and $f_a$ be a function with the same domain and codomain. The operations are defined as follows:

\begin{itemize}
\item Abstraction: $\lambda(R_f, R_a) = R$, such that $R(a_i, v_j) = R_f(a_i, v_j) \lor R_a(a_i,v_j)$ for all $a_i \in A$ and $v_j \in V$ --i.e., $\lambda(R_f, R_a) = R_f \cup R_a$.
\item Containment: $\eta(R_a, R_f)$ is true if $R_a(a_i,v_j) \to R_f(a_i,v_j)$ for all $a_i \in A$ and $v_j \in V$ (i.e., material implication), and false otherwise.
\item Reduction: $\beta(f_a, R_f) = f_v$ such that, if $\eta(f_a,R_f)$ holds, $f_v(a_i) = R_f(a_i)$ for all $a_i$, where the random distribution is centered around $f_a(a_i)$. If $\eta(f_a,R_f)$ does not hold, $\beta(f_a, R_f)$ is undefined --i.e., $f_v(a_i)$ is undefined-- for all $a_i$.
\end{itemize}
 
Abstraction is a construction operation that produces the union of two relations. A function is a relation and can be an input to the abstraction operation. Any relation can be constructed out of the incremental abstraction of an appropriate  set of functions. The construction can be pictured graphically by overlapping the graphical representation of the included functions on an empty table, such that the columns correspond to the arguments, the rows to the values and the functional relation is depicted by a mark in the intersecting cells. 

The containment operation verifies whether all the values associated to an argument $a_i$ of $R_a$ are associated to the same argument of $R_f$ for all the arguments, such that $R_a \subseteq R_f$. More generally, the containment relation is false only in case $R_a(a_i,v_j)=1$ and $R_f(a_i,v_j)=0$ --or if $R_a(a_i,v_j) > R_f(a_i,v_j)$-- for at least one $(a_i,v_j)$. 

The set of functions that are contained in a relation, which is referred to here as the \emph{constituent functions}, may be larger than the set used in its construction. The constituent functions are the combinations that can be formed by taking one value among the ones that the relation assigns to an argument, for all the arguments. The table format allows to perform the abstraction operation by direct manipulation and the containment test by inspection. The construction consists on forming a function by taking a value corresponding to a marked cell of each column, for all values and for all columns. The containment is carried on by verifying whether the table representing the function is contained within the table representing the relation by testing all the corresponding cells through material implication.

For this, the abstraction operation and the containment test are productive. This is analogous to the generalization power of standard supervised machine learning algorithms that recognize not only the objects included in the training set but also other objects that are similar enough to the objects in such set.

Reduction is the functional application operation. If the argument function $f_a$ is contained in the relation $R_f$, reduction generates a new function such that the value assigned to each of its arguments is selected randomly from the values assigned to the same argument in $R_f$. If $f_a$ is not contained in $R_f$ the value of such functional application operation is not defined. The reduction operation selects a function included in the relation on the basis of a ``cue'' function. The lower the entropy of the relation the larger the similarity between the object retrieved from the relation and the cue. In the limiting case, when the entropy of the relation is zero but reduction is applicable, the reduction selects the cue itself.

The distinction between non-entropic and entropic computing corresponds to the contrast between  ``local'' versus ``distributed'' representations. According to Hinton \cite{Hinton-1986}, TMs hold local representations in which the units of form are related \emph{one-to-one} to their corresponding objects at the level of meaning or content, while in distributed representations this relation is \emph{many-to-many}. Representations in Turing machine are strings of symbols on the tape where each word denotes a particular basic unit of content and non-basic meanings are produced by the composition of the meanings of words into meanings of phrases and sentences, but basic words are never overlapped on the tape or the memory, and representations are local. In this setting basic representational objects are mutually independent and the entropy is zero. This contrasts with the representation of a set of functions overlapped on a table, where a marked cell can contribute to the representation of more than one function and a function can share marked cells with other functions. Hence, the representation is distributed and the entropy measures its indeterminacy.

Standard TMs can compute relations by computing its constituent functions. However, the information provided by the intersections is lost if the functions are considered as independent objects. The computing entropy measures the information provided by such intersections. If the functions constituting a relation are mostly independent, there are few intersections and the entropy is high; however, if there is a large number intersections the entropy is decreased and computations become more determined accordingly. 

The inclusion of the entropy in a theory of computing gives rise to a new computational trade-off that here is called \emph{The Entropy Trade-Off}: If the entropy of the machine is very low the computation is pretty determined, being the TM the extreme fully determinate case whose entropy is zero; in the other extreme, if the entropy is too high, the information is confused and computations are not feasible. However, if the entropy is moderate there is a certain amount of indeterminacy in the computing engine, that allows multiple behaviors and computations are still feasible. 

The entropy trade-off of computing engines corresponds to the potential productivity of decisions in the environment. 

\begin{figure}[t]
\includegraphics[width=0.8\textwidth]{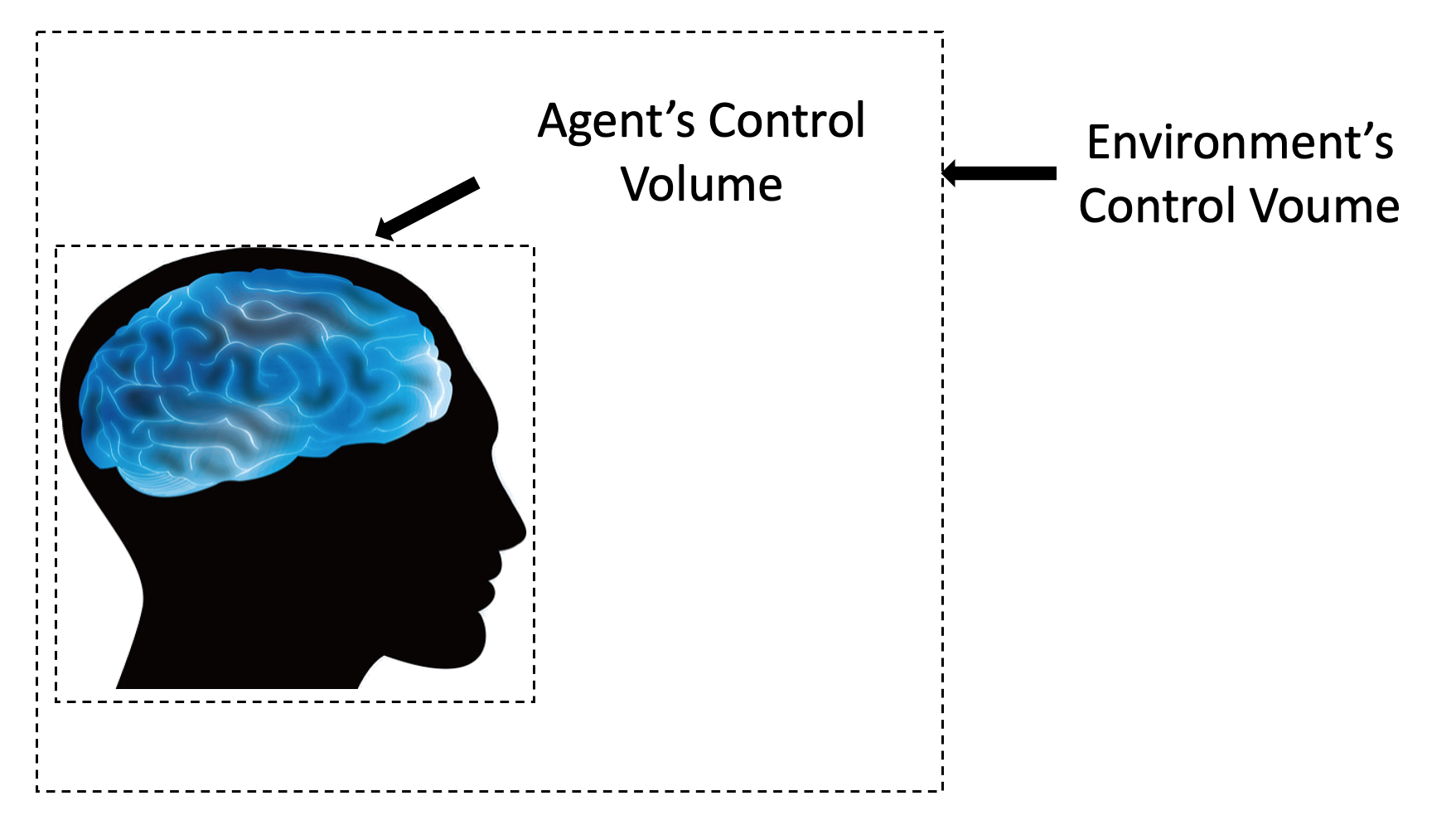}
\centering
\caption{Embedded Control Volumes}
\label{fig:control-volume}
\end{figure}

The interaction of the agent with the environment introduces some indeterminacy to the computing process. The external input, such that the face resulting of flipping a coin, is input into the agent's control volume, as illustrated in Figure \ref{fig:control-volume}, and from its point of view the value of such variable is genuinely random. Control volumes can be embedded in more extensive environments from the most specific ones to the universe as a whole. Whether spontaneous events can occur within a control volume or the knowledge of such events is a case of incomplete information is the deep question whose answer opposes determinism and the indeterminate views of the world.

\section{Table Computing and Associative Memory}
\label{sec:tables}

The implementation of RIC in a table format is referred to as \emph{Table Computing} \cite{Pineda-modo-comp}. This format was used to define and implement an associative memory system, where standard tables are used as \emph{Associative Memory Registers} (AMRs) \cite{pineda-entropic-mem-2020}. The system uses three main algorithms that perform operations between the corresponding cells of memory registers directly, which are called here $Memory\_Register$, $Memory\_Recognize$ and $Memory\_Retrieve$. These algorithms implement the $\lambda$, $\eta$ and $\beta$ operations respectively, and are so simple that here are called \emph{Minimal Algorithms}.
The objects stored in the AMRs are distributed representations of individuals or classes of individuals. These representations have an entropy level depending on the information stored in the AMRs at a given time, and the memory system conforms to the entropy trade-off.  The memory operations are computed in parallel, both between the corresponding cells of AMRs and between the full set of AMRs included in the memory system. This parallelism is a property of the memory at the computational system level in Marr's  sense \cite{Marr} and not only a contingency of the implementation. The power of the distributed representation comes from the coordinated simultaneous computing at all local memory cells and the AMRs.

\subsection{A visual memory for hand written digits}
\label{sec:experiment}

The associative memory system was used for storing distributed representations of hand written digits \cite{pineda-entropic-mem-2020}. In the experiment an associative register for holding the representation of each one of the ten digits was defined. All instance digits in the training corpus were placed on an input visual buffer, and mapped into a set of features with their corresponding values  --which is called here \emph{the space of characteristics}-- through a standard deep-neural convolutional network \cite{Lecun-nature}. Intuitively, this corresponds to seeing each digit, mapping it into its corresponding abstract modality independent representation through a bottom-up perceptual operation, and registering it into its corresponding associative register through the $Memory\_Register$ algorithm. 

The memory recognition operation was implemented by mapping the digit to be recognized into its corresponding representation through the same deep-neural network, and applying the $Memory\_Recognize$ algorithm. 

The results show that if the entropy is zero or very low the overall recognition precision and recall are very low; that both precision and recall increase according to the increase of the entropy, but that recall decreases when the entropy is very high; that there is an interval of moderate entropy values in which precision and recall are both very satisfactory; and that the entropy trade-off holds. Such entropy interval determines the operational characteristics of the associative memory system.

The experiment of memory retrieve consists on presenting arbitrary hand written digits to the memory to be used as cues and map the recovered function in the space of characteristics into a concrete representation in an output visual buffer through transposed convolution, as in standard architectures to generate pictures out of abstract characteristics within the deep neural networks framework \cite{radford-2015}. The results show that the objects retrieved from memory are similar to the cue only when the entropy is very low, and that the similarity decreases quite sharply with small entropy increments. Memory retrieve is a constructive operation such that the retrieved objects go from ``photographic'' images, to similar images, to imaged objects, depending on whether the entropy is zero, moderate or too high respectively.

The overall functionality of the memory system conforms to the intuition that memory recognition is a flexible and robust capability but recovering or retrieving objects from memory is a much more selective and restrictive operation.

In the present system information is stored and recognized locally through the basic logical disjunction and material implication operations between cells of tables or associative registers, in addition of the standard assignment operator and a random generator. The memory system is associative, distributed and declarative, and has a dual symbolic and sub-symbolic interpretation. The experiment was implemented as a simulation in a standard processor with a graphic GPU board, but the hardware construction of the device should not be problematic with current integrated circuits technology.

\section{Entropy and Cognitive Architecture}
\label{sec:cog-arch}

\begin{figure}
\includegraphics[width=1\textwidth]{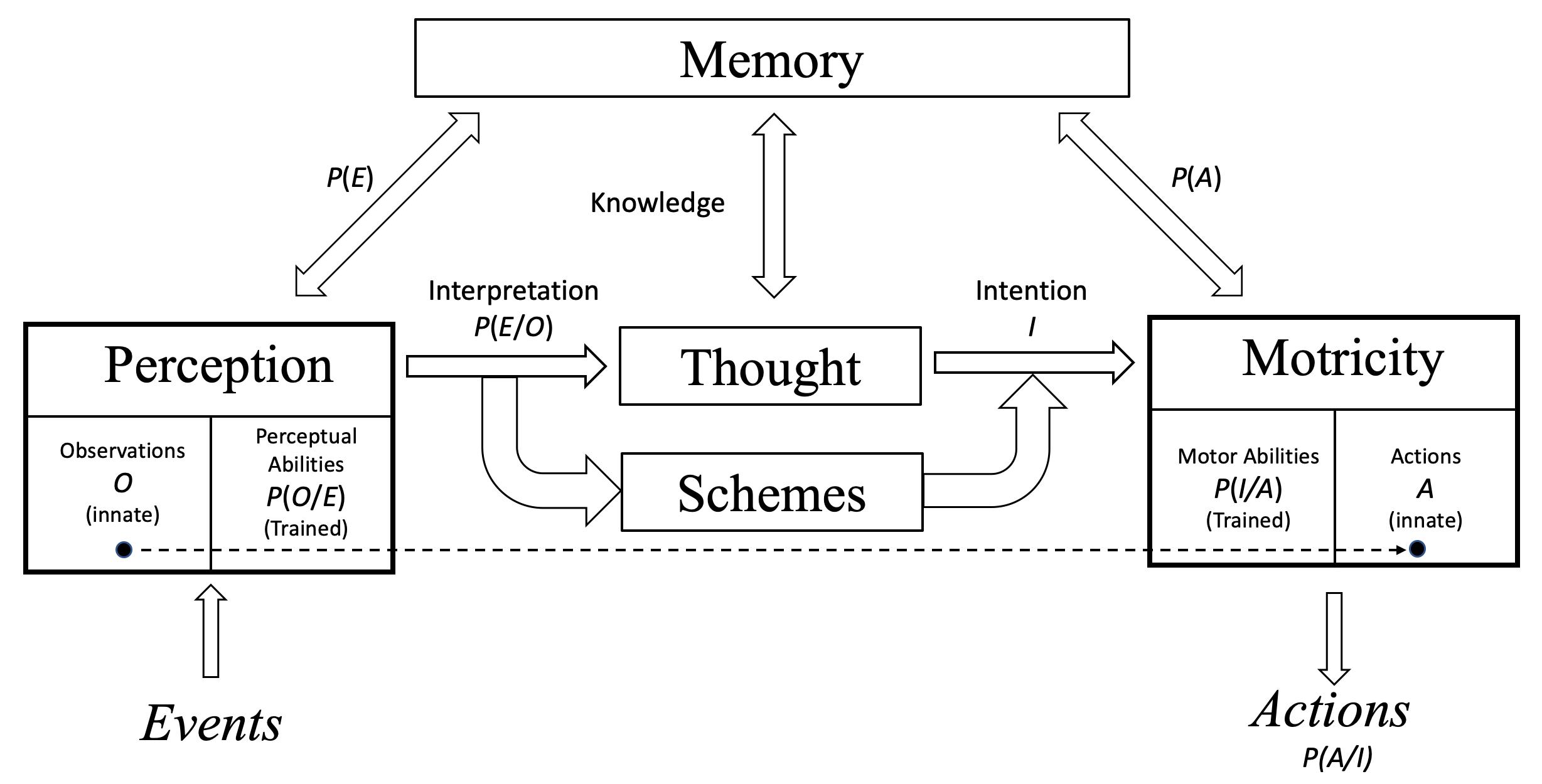}
\centering
\caption{Cognitive Architecture}
\label{fig:cog-arch}
\end{figure}

The computing engine that is causal and essential to cognition and rationality needs to be related to the cognitive architecture of the agent. An architecture including perception, motricity or action, thought, schematic thinking and reactive behavior, which here is called \emph{Interaction-Oriented Cognitive Architecture} or \emph{IOCA},\footnote{IOCA is an abstract generalization of the architecture with the same name that was implemented in the service robot Golem-III \cite{golem-III-2020}. Publications and videos about this robot are available at \url{http://golem.iimas.unam.mx}. Golem-III was built with standard digital computers and its only source of entropy is the external environment, which is very predictable in the traditional competition settings and benchmarks for this kind of devices, and the overall behavior, including the decisions made by the robot, is quite predetermined.} is illustrated in Figure \ref{fig:cog-arch}. The architecture is stated in terms of functional modules, in the sense of other cognitive architectures such as SOAR \cite{soar-2012} and ACT-R  \cite{anderson-2004}. The modules are stated at the computational theory in Marr's sense \cite{Marr} and hence independently of particular representational formats or processing strategies. The purpose of this presentation is to illustrate the roles of the entropy, the potential productivity of decisions and the relational-indeterminate computing in cognition.

Decision making is at the center of thought and rationality. Theories of rationality may be classified in three main kinds: i) those that support a general mechanism that works according to first principles; ii) those that support that thinking is schematic and carried out through a rich variety of specific mechanisms that allow an effective interaction with the environment; and iii) those that support both modes. The first and second kinds are depicted by the modules \emph{Thought} and \emph{Schemes} in Figure \ref{fig:cog-arch}. The present architecture is an instance of the third kind.

Theories sustaining that thought uses a general mechanism and conforms to first principles are exemplified by the logical and propositional view of knowledge; by the bayesian reasoning approach to induce causes of events or phenomena on the basis of their observable effects; and by theories holding that decision making is a matter of maximizing the expected value of decisions and actions, as in game theory.

Thought was addressed since the origins of Artificial Intelligence and the proposal of \emph{Bounded Rationality} \cite{simon-1957,bounded-rationality}. The original program used the minimax decision making strategy although computed with limited computational resources and aided by heuristics. In this respect it departed explicitly from the omniscient rationality and from the notion of economical or administrative man advocated by Von Neumann and Morgenstern in \emph{Game Theory and Economic Behavior} \cite{von-neumann-morgestern}. This research program gave rise to the \emph{Physical Symbol Systems Hypothesis} holding that a system of grounded symbols provides the necessary and sufficient conditions to produce general intelligence \cite{Newell-Simon,simon-1996}, and was further developed with Newell's claim that there is a system level directly above the symbol level, that he called \emph{the knowledge level}, in which the only rule of behavior is what he called \emph{The Principle of Rationality} \cite{Newell}. 

This program was later developed into the SOAR \cite{soar-2012} and ACT-R  \cite{anderson-2004} cognitive architectures, the former focused on modeling AI tasks and applications and the latter on developing a theory of mind and its relation to the brain. In this paradigm, decision making was conceived as a pure thought process where rationality was enacted through symbolic manipulation; interpretations were already available in the symbolic format; and the actions performed with the purpose of satisfying the agent's decisions achieved their intended effect necessarily. In practice, interpretations and actions were performed by human-users and computational thought was detached from the world, as in computer chess and similar programs.

This limitation of the original program of AI has been addressed from different perspectives such as neural networks \cite{Rumelhart} and probabilistic causal models (e.g., \cite{pearl-2009,sucar-2015}), which reject symbolic representations, and embedded architectures \cite{Brooks}, embodied cognition \cite{Anderson-b:2003} and enactivism \cite{Froese}, that reject representations and symbolic computing altogether. However, regardless whether or not thought and memory are representational or implemented through symbolic or sub-symbolic computing, these faculties have to be connected with the world through perception and action in a principled way.

Schematic thinking, on its part, consists on mapping interpretation into actions through a specialized module. Schemes may be innate, although they may be used contingently by analogy to learned concepts, or may be developed through experience and manifested as habitual behavior. The \emph{Schemes} module is exemplified by the so-called procedural representations in Artificial Intelligence, such as Minsky's Frames \cite{minsky-1986}, that opposed production systems and logical approaches, which may be considered general reasoning mechanisms.

Theories of thought and decision making should also be seen in relation of two perspectives: the mechanisms that the agent uses to make decisions and the constraints of the environment that allow such decisions be enacted in the world. The potential productivity of decisions measures the overall impact that decisions can have in an environment, but particular decisions should take into account such constraints and this knowledge should be available in memory.

Since the early work of Simon \cite{simon-1955} theories of bounded rationality distinguish behavioral constraints that relate to the decision maker, from external constraints that refer to the ecological structure of the environment. However, the characterization of what is ``behavioral'' and what is ``external'' is by no means easy to make. In practice, reasoning has traditionally lean to model behavioral constraints and the need for an ecologically motivated research program has been called for (e.g., \cite{koehler-1996,cosmides-1996}). This concern has been addressed more recently within the so-called \emph{Ecological Rationality} (e.g., \cite{todd-gigerenzer-2012}) which investigates the coupling between the heuristics and the features of the environment for acting effectively in the world through a continuous interacting cycle. Heuristics in this latter paradigm are commonly described as algorithmic encapsulated processes.

The dual theory, in turn, considers that thinking and problem solving do require a general deliberative mechanism, but thought is an expensive resource and this module may be bypassed by schemes that map the output of perception into the input of the motricity module. Conversely, schemes may be quite rigid and lead to irrational behavior, but this can be avoided if the agent is a competent subject and has the interest and the energy to address the task through deliberative thinking (e.g., \cite{barbey-sloman-2007}).

The observations can also be connected directly with basic actions rendering reactive behavior, as illustrated by the dotted arrow in Figure \ref{fig:cog-arch}. In this architecture the agent is always engaged in a cycle of perception, schematic behavior and action, that uses thought on demand and embeds reactive behavior; hence the name IOCA. The architecture is only a schematic idealization, as in natural cognitive architectures the functional modules may overlap and the boundaries between them may be loosely demarcated.

\subsection{Interpretation and Action}
\label{sec:interpret-action}

The inputs of the perception module are the observations that the agent sense in the environment and the relevant knowledge stored in memory, and its outputs are the corresponding interpretations. These are in turn the inputs to the thought module, in which reasoning is performed. Thought interacts with memory, where knowledge is registered and recalled, and its outputs are the intentions of the agent, which are construed here as specifications for actions that can be rendered through linguistic or motor behavior. The motricity module has as its inputs the intentions produced by thought but also interacts with memory, and its outputs are the actions that the agent performs in the environment. 

The perception module is stated in the Bayesian perspective. The agent is endowed with the capacity to make observations $O$, by organizing the information provided by the senses into a basic entity, property or scene, such as a physician feeling the heat of the patient's body or the chess player identifying the pieces on the board. The observations can be performed by anyone who has not a particular impairment preventing him or her to do so, and can be considered innate. Observations are the manifestation of events in the world, and perception produces the hypotheses of what are such events on the basis of the observations and the relevant knowledge. The term $P(E/O)$ stands for the most likely event that happened in the world given the observation. As there may be many potential events, the best hypothesis is the event that maximizes this value for the given observation. This term is the output of perception or \emph{The Interpretation}.

This is computed by Bayes's theorem which is stated as follows: $P(E/O) = ArgMax_EP(O/E) \times P(E)/P(O)$. The term $P(O/E)$ --the \emph{likelihood}-- represents, for instance, the ability of a physician to be able to tell how likely is that the patient has fiver if he or she has typhoid --the observation and the event respectively. This term represents the expertise of the physician that is acquired through the years of practice, and is readily available when an observation is made. Here this term is referred to as \emph{The Perceptual Ability}.

The term $P(E)$, on its part, usually referred to as \emph{the prior}, stands for the knowledge of the agent of how likely is that such an event occurs in the world. For instance, the physician may know that 9 out of 10 children in town have typhoid. 
 
Finally, the term $P(O)$ is the probability of the observation. This probability is in turn $P(O/E) \times P(E) + P(O/\overline{E}) \times P(\overline{E})$ where the first and second terms of the sum correspond to the true and false positives --the times the observation was produced by the event and the times the observation was produced by other events. However, the term $P(O)$ is dropped in the standard formulation of the noisy channel because the computation aims to select the most likely event in relation to the same observation, and the full expression is simplified to $P(E/O) = ArgMax_EP(O/E) \times P(E)$. The resulting value is a weighing factor rather than a probability, but it is enough to compute the best interpretation. The bayesian expression states simply that the best interpretation hypothesis is that the event that most likely happened is the one that maximizes the product of the likelihood and the prior. 

The hypothesis that people make interpretations using Bayes' theorem has been tested and the early empirical evidence suggested that this is not the case \cite{tversky-1974,casscells-1978} and that people normally ignore the priors when individuating or specific information is available, the so-called base-rate neglect or fallacy (e.g., \cite{tversky-1974}). However, base rates interact with specific information, and whether they are considered depends on their relevance for the task at hand \cite{bar-hillel-1980}. More recently it has been argued that the experiments that gave rise to such result were based on the implicit assumption that people, for instance, physicians, use Bayes' theorem in the standard probability format. In those experimentes the posterior probability $P(E/O)$ had to be computed given the prior, the likelihood and the rate of false positives (often called the base rate, the hit rate and the false alarm rate) and the experiments assumed that the Bayes' theorem is known and can be used operationally, but as normal people, such as most physicians, are never taught probability theory and are unfamiliar with the probability format, this knowledge must be innate and subconscious. No surprisingly people in general do very poorly. 

However, presenting the information as natural frequencies do support natural bayesian reasoning \cite{gigerenzer-1995,cosmides-1996} and that base-rate neglect is unfounded \cite{koehler-1996}. For instance, if instead of the actual probabilities a physician knows the number of people in his or her demarcation, those who have presented an illness and had the symptoms, and those who had the symptoms but did not have the illness, he or she can compute the actual posterior probability, although using a simpler but correct form of Bayes' formula. Gigerenzer \cite{gigerenzer-1995} calls the latter form of acquiring the information the \emph{natural frequency format} and argues very clearly that the same mathematical object --in this case Bayes' theorem-- can have different representations and be computed by different algorithms, and that the information should be presented in the appropriate format.

From an ecological and evolutive perspective the information format must be available directly in the environment (e.g., \cite{cosmides-1996}) but the interpretation machine must map such format into one akin for memory and processing too. Clearly, if the information is presented in the probabilities format but the processor and memory use a natural frequencies format or vice versa, there would have to be a costly and unnatural translation.
 
An instance of the natural frequencies format \emph{all the way through} is illustrated by the associative memory presented in Section \ref{sec:tables}. There, the digits instances in the external medium are feed in serially into their corresponding associative registers through a transfer function, implemented by the neural network, that places the information in the standard configuration of the table computing machine. This format consists on a large number of abstract amodal features that are analogous to their modality specific input and output representations --the digits on a piece of paper or a computer screen that are input and output through modality specific buffers-- and all instances of the same digit are represented in a table but abstracted through the logical disjunction operation. Hence, the format used by the memory corresponds to a natural frequencies format.

The mapping from the input modal buffer into its corresponding amodal representation corresponds to computing a likelihood; recognizing or retrieving the content of the memory registers corresponds to computing a prior; and the memory recognition and recall operations correspond to selecting the object that matches the cue from its right memory register among all the other registers, and ``maximizes''  ``the product''  of ``the likelihood'' and ``the prior''. For instance, in the experiment in Section \ref{sec:experiment} the likelihood is the descriptor representing the cue provided by the input neural network --which plays the role of the bottom-up or low-level perception; the ``events'' are the digits presented to the agent for recognition or recall, whose representations are stored in their corresponding associative memory registers. These contain the abstractions of all instances of the corresponding digits previously seen and registered; and the maximization operation corresponds to recognizing and retrieving one specific digit among the ten possible ones, on the basis of the cue.

This suggests that Bayesian interpretation does not mean necessarily that people actually compute the posterior probability and that there may be other modes of computing that implement the process in a more effective way. The purpose of the computation is to interpret the observation presented for recognition: a clinical physician is concerned with choosing the most likely disease given the symptoms and his or her knowledge and experience, in order to decide the best treatment, but he or she does not need to know mathematics and make the actual computations, as it is the case commonly.

From the neurosciences perspective, perception, action, learning and memory accord to Bayesian principles, the so-called Bayesian brain hypothesis \cite{Friston-2010}. The implicit Bayesian maximization involved in the production of the interpretation may require a cycle of memory recall operations such the current interpretation is compared with the likelihood of the next observation, and the final interpretation is stable enough to be reliable.

There may be a wide variety formats, representational systems and processing strategies and the basic intuition underlying Bayes' theorem can be generalized into the proposition that the best interpretation strategy is to select the best hypothesis that results from pondering the information provided by the perceptual abilities with the relevant knowledge available in memory. This intuition is very strong and is referred to here as \emph{The Principle of Interpretation}.

The Bayesian analysis can be applied for the action part or the motricity module that generates the action, as illustrated in Figure \ref{fig:cog-arch}. In this case the inputs are the intentions $I$ produced by thought which need to be enacted through the actions that the agent can perform, and the output is the best action that achieves the intention, which is designated by the term  $P(A/I)$. The bayesian expression for the action case is $P(A/I) = ArgMax_AP(I/A) \times P(A)$. As in the interpretation case, the goal is to chose the best action in relation to the same intention, and the probability of the intention can be dropped.

The likelihood $P(I/A)$ represents the extent to which performing a particular action renders the given intention, or the extent to which the intention can be achieved given that the action is performed. For instance, that a melody will be produced given that a sequence of keys is pressed on the piano. The primitive or innate actions $A$ can be performed by anyone, such as pressing the keys, but to produce music by playing the piano requires years of rehearsing. The ability is acquired through practice and experience, as in perception, and this latter likelihood is called here \emph{The Motor Ability}. 

The term $P(A)$, on its part, represents the feasibility that the action can be enacted in the environment. This involves behavioral constraints of the agent, such as his or her capacity to perform such action, or the cost that needs to be afforded for achieving it, but also external aspects that depend on both the physical environment and the dispositions and intentions of other agents or the society. The external aspects correspond to the potential productivity of decisions, but for particular actions. If this knowledge is not considered the actions may not achieve the intended effect due to factors that are not under the control of the agent.

Finally, the bayesian law selects the action that maximizes the product of the motor abilities with the cost or feasibility of the action. As in the interpretation case, the motor ability and the knowledge about actions can be expressed through a variety of formats and computing strategies.

On the basis of these consideration the bayesian law can be generalized into the proposition that the best strategy for acting in the world is to select the best action that satisfy the intention by pondering the information provided by the motor abilities with the knowledge about the actions available in memory. This latter proposition is referred to here as \emph{The Principle of Action}.

Perception and action are commonly modeled through standard algorithms and the entropy of the machine is not considered or is zero. However, natural memory is entropic most likely, memory recall is an indeterminate operation, the links from memory into perception and action introduce a level of indeterminacy, and perception and action conform to the entropy trade-off.

\subsection{Schematic Thinking}
\label{sec:schematic-thinking}

Schematic behavior relates perception and action through specific highly specialized modules. A paradigmatic form of schematic behavior may be daily-life intentional behavior --walking, eating, bathing, talking, etc. These kinds of actions bypass thought and relate perceptual and motor abilities directly, as long as the expectations of the agent are met in the world. However, when a spontaneous event occurs in a relevant respect, there is an interruption and rational agents engage in a deliberative thought process. The schematic behavior may be put on hold to attend the event, or may be continued and performed simultaneously with thought, but the attention is focused on this latter process.

In Cognitive Psychology schematic thinking is exemplified by the use of heuristics that explain biased but systematic behaviors \cite{tversky-1974}, that were presented in opposition to Bayesian Inference \cite{gigerenzer-1995,cosmides-1996}. In such view the heuristics handle diverse sorts of common interpretation rightly and very effectively, although on occasions they can lead to irrational behavior.

A more recent illustration provided within the ecological rationality is the so-called gaze-heuristics which is used to track objects following a trajectory, such as baseball players catching a ball. It consists on fixating the gaze on the ball, start running and adjust the speed so that the angle of gaze in relation to the ground remains constant  \cite{todd-gigerenzer-2012}. In this behavior there is a continuous interpretation and action process, and the essence of the heuristics is to compute the speed as a function of the angle. In a physical model the full trajectory would have to be computed, but, according to ecological rationality, this is too complex and people are unable to do it. Nevertheless, in such view it is held at the same time that the angle and the speed are much simpler computations that are actually performed by people in real time. However, computing these parameter requieres significant computational resources, metric systems and measuring devices, which might not be available to people and other higher evolved animals that can achieve these tricks; and natural computing should proceed by other means.

In the present proposal computing the scheme can be construed as follows: the perceptual ability produces the angle --represented in a space of abstract and amodal characteristics-- out of the image in the input visual buffer; this interpretation is the argument of a scheme mapping the angle into a speed --in the same abstract and amodal space--  but implemented through table computing or some form of analogical or diagrammatic reasoning, which does not use a standard algorithm involving costly computations. The speed is in turn the input to the perceptual ability and the motricity module renders the action through the motor actuators. 

This basic model does not resort to memory and although computing the scheme directly may be very effective, the agent may not be able to adapt to even slight changes in the environment. A more robust model would include a perceptual and a motor memory for enriching the input interpretation and considering the potential contingencies of the action, and the overall behavior would be informed by the knowledge of the agent, according to the full use of the principles of interpretation and action. 

Schemes compute functions or relations whose arguments and values stand for the interpretations and intentions respectively. Schemes are normally implemented through standard algorithms and compute by TMs, but they can be implemented by other means, such as neural networks, analogical and diagrammatic machines, or the relational-indeterminate implemented through table computing, among other specialized modes of computing. In case entropic processes are included, behavior would be more flexible and should obey the entropy-trade off.

In any case, motor actions produced through schematic behavior are highly conditioned by perception and the scheme proper. To say that the baseball player makes the decision to increase or decrease its speed and the positions of his or her body organs is a manner of speaking. Explicit decision making and deliberative thought allows people to anticipate the world in the short, middle and long term, and the relation between interpretation and intentional action is mediated by knowledge and values. For this, deliberative and schematic thinking should be distinguished.

\subsection{Thought and Decision Making}
\label{sec:decision-making}

\begin{figure}
\includegraphics[width=1\textwidth]{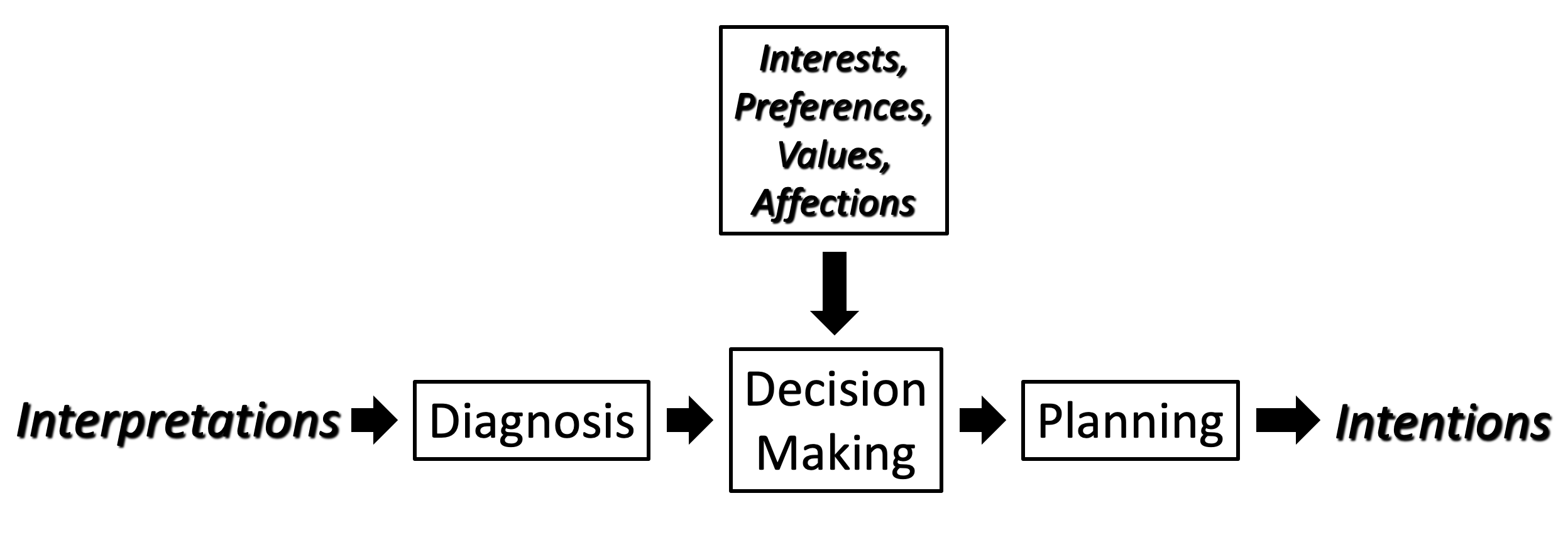}
\centering
\caption{The Inference Pipe-line of Thought}
\label{fig:thought}
\end{figure}

The thought module is zoomed in Figure \ref{fig:thought}. Its inputs and outputs are the interpretations produced by perception and the intentions that need to be enacted by the motricity module respectively. The thought process consists on a pipeline of a diagnosis, a decision making and a planning inferences. 

The input to the diagnosis module is the interpretation of the event that caused the interruption of the schematic cycle and its output is a hypothesis of its cause, which is the main input to the decision making module. This latter module has other inputs such as the agent's interest, preferences, values, affections, etc., which may have a subjective component, and its output is the decision proper: what to do about the interpretation hypothesis. The decision is in turn the goal of the planning module whose output is the specification of the actions that need to be carried out in order to enact the decision and the output of the thought process as a whole is the intention. The intentions are enacted as actions and the agent must verify that the effects of the actions are as expected and may remain engaged in an inference cycle until the intended effects are achieved. This cycle is also accompanied by learning, so the agent is better prepared to deal with similar contingencies in the future.

For instance, someone enters into his or her house and sees a puddle in the living room coming out of the kitchen. This is already an interpretation that can be express linguistically. The problem for the house owner is what to do about it but first he or she needs to find out what is the cause of the observation, and a diagnosis inference is performed. This process can be thought of in bayesian terms but also as a case of abductive symbolic reasoning, which underlies the same basic intuition. The diagnosis involves the synthesis of a set of hypotheses, such as that a pipe was broken or that his or her child forgot to turn off the faucet. The output of the diagnostics module is a ranked set of hypothetical causes, which is the main input to the decision making module, in addition to the subjective inputs as discussed above and illustrated in Figure \ref{fig:thought}, and its output is the decision proper. If the diagnosis was that the sink tap was turned on, the decision is to turn it off and dry the puddle; but if the diagnosis was that a pipe was broken, the decision is to fix it. Another potential decision is doing nothing, but the subjective inputs, such as the interests and values, place the decision in a deeper perspective, and modulate the decisions that particular agents make.

The decision becomes the goal of a plan, that has to be induced and executed. For instance, if the decision is to turn off the tap and dry the puddle, the plan may be go into the kitchen, stand in front of the sink, turn off the tap, get a mop and dry the floor; but if it is to fix the pipe, the plan may be turn off the house main water valve and phone the plumber. The output of the planning module is the specification of the actions that need to be performed with the purpose that the decision is enacted. This specification is the intention, which is the input to the motricity module. Some of the actions may be performed linguistically and other by physical motor behavior, but in the end these are all motor actions. The agent needs to monitor whether the effects of the actions are as expected through the main interaction cycle with the world, and remain engaged recurrently until the plan is achieved and the problem that gave rise to the thought process is fixed.\footnote{This cycle of inference has been implemented in the robot Golem-III. Videos showing the robot performing simple tasks but involving the full inferential pipe are available at \url{http://golem.iimas.unam.mx/inference-in-service-robots}}

An additional input to the decision making module is the knowledge whether the decision can be enacted in the world. Some of this information depends on the types of actions that the agent can perform through its motricity module, but there is an additional dimension that depends on the knowledge of the causal structure and uncertainty of the world. Each particular decision made by a particular agent may or may not be enacted due to environmental or ecological aspects, and a good decision maker should take into account such external factors. 

More generally, decision making is placed within the two different sources of indeterminacy: the computing entropy of the engine that carries on with the thought process, and the entropy of the environment, and there is a \emph{Decision Making Trade-off}: the best decision is the one that results from pondering the value of the potential decisions with the uncertainty that they can be enacted in the environment. The largest the value of the decision for the agent and the lesser the uncertainty that such a decision can be enacted, the better the decision. This is also an application of the Bayesian principle, where the value of the decisions to achieve and intended effect can be seen as a likelihood and the knowledge of whether the effect can be enacted in the world as a prior.

As in perception, action and schematic behavior, it is possible to hypothesize that deliberative thought is performed through direct mechanisms and has an implicit entropy, and conforms to the entropy trade-off.

\subsection{Memory versus Abilities}
\label{sec:memory}

Knowledge stored in memory opposes perceptual and motor abilities in several dimensions. While the former is available in memory, can be expressed declaratively, and is acquired and learned through language, the latter are deeply embedded in the perceptual and motor structures and are acquired through extensive training. While knowledge is registered, recognized and retrieved from memory and used in language, abilities are used but the information that is causal to their deployment cannot be recalled or remembered. While  knowledge is transparent to consciousness, at least when is stored or is retrieved from memory or used in reasoning, and can be the object of reflection and introspection, the experience of deploying an ability is felt but opaque to consciousness; and while knowledge is causal and essential to intentional behavior, the reports or explanations of what is the ``knowledge'' involved in abilities are a posteriori reconstructions of the processes that are causal to the experience itself. Abilities are embedded deeply into the perceptual and motor engines, but knowledge is better thought of as the object that is held in memory. Schemes are similar in most of these respects to abilities, although may have a larger innate component, and oppose to knowledge in the same dimensions. 

In AI symbolic systems knowledge is held in knowledge bases, that may hold incomplete information, may involve interpretation heuristics and may perform conceptual reasoning directly, and have a mostly symbolic character. Knowledge can be stored and retrieved, used in reasoning and inform perception and action, and the functional module that holds such a kind of information may be considered a memory proper. 

Symbolic knowledge-bases oppose sub-symbolic machines, such as neural networks. These latter machines cannot express symbolic structures; hence, cannot hold information declaratively, and individual or episodic memories cannot be recalled. For these reasons sub-symbolic systems are better thought of as transfer functions for classification or prediction, for instance, but should not be considered proper memories (e.g., \cite{Fodor-Pylyshyn}).

% For instance, a bank's customers or a taxes data-base cannot be defined in these latter formalisms, because the information of customers and taxpayers would be confused and could not be recovered.

The associative memory in the present architecture has both a symbolic and a sub-symbolic aspects. The interpretation of the register's content as a whole has a symbolic character but its structure is sub-symbolic. Memories are entropic computing machines and the entropy trade-off establishes their operational range. The associative memory registers in the experiment reported in Section \ref{sec:experiment} hold only basic units of content, such as the digits. This is a proof on concept experiment, and it is an open question whether other kinds of individual objects can be modeled, or whether larger structures for holding composite contents can be created, but the experiment suggests that the causal and essential engine that gives rise to memory may have such a kind of distributed structure, that hold the information in natural but abstract formats, and computing is performed through minimal algorithms. 

The cognitive architecture also suggests the possibility that logical reasoning can be a memory operation in which the premises and conclusions of arguments are represented in the space of characteristics, such that the representation of the premises is included in the representation of the conclusion. This would allow for a representational view of knowledge, but one in which the actual symbols appear only in the input and output modality specific buffers, and the ``symbolic manipulation'' consists on memory operations in the abstract characteristics space.

% It is feasible that during training some knowledge in memory is transferred into the abilities, and that some information gained through the abilities is ``rationalized'' and expressed linguistically, and there may be transfers from memory to perception and vice versa. This distinction is however schematic and there are objects, such as feelings and affections, that may have both components.

\subsection{Natural Computing and Minimal Algorithms}
\label{sec:schemes}

Natural frequencies do present the information in a manner more akin to the natural mode of computing than other formats, such as standard mathematical notation. However, according to standard thinking in Cognitive Science, computing is nevertheless performed using standard algorithms in a TM or some equivalent machine. The intuition is that the human brain is such a powerful machine that can easily evaluate the arithmetic operations involved in the computation (e.g., \cite{cosmides-1996,todd-gigerenzer-2012}); hence people do use algorithms in the manner that they are computed by digital computers of the standard sort. 

However, the format of the Turing Machine is linguistic and propositional --representations are strings of symbols on the tape-- and if the brain were a TM the natural frequency format would have to be translated into the symbolic or propositional one, and computing would be equally hard. Conversely, the probabilistic format is a propositional representation and if this were the actual format employed by the machine, probabilistic reasoning in this format would be easy. More generally, for computing to be effective the format in which the information is presented should be alike to the actual format in which the computations are performed.

This proposition is stated in more general terms in the theory of Turing Machines and computability, and is a fundamental concept in computer science and the construction of computing machinery: TMs' representations are subject to a set of \emph{interpretations conventions} and to a \emph{standard configuration}. The former are needed to interpret the workings of the machine, starting with the most basic convention that a TM computes a function, and that the input and output strings represent the argument and value respectively. 

%The interpretation conventions include also the notation -for instance, the string ``111'' stands for three in monadic notation, for seven in binary notation and for one hundred and eleven in decimal notation.

The \emph{standard configuration} in turn specifies the format of the representations in relation to the finite control including the scanning device and the computing medium; for instance, if the medium is a tape, a grid or a set RAM registers. The configuration specifies as well how the input and output devices, such as the keyboard and the monitor, must place and take the information from the central processing unit, and also allows that the output string of one computation can be the input to the next. The interpretation conventions and standard configuration of table computing described in Section \ref{sec:tables} are specified according to these criteria.

%There may be several algorithms that compute the same function and the same algorithm may compute a different function under different interpretation and/or configuration, and there may several modes of computing that evaluate the function by other means \cite{Pineda-modo-comp}.

%In the so-called input and output standard configurations it is required that the scanning device is placed on top of the cell of the tape including the leftmost symbol of the input and output strings when the computation started and ended respectively, and that all cells to the left and right of the string on the tape are blanks in these two states. 

%Another basic interpretation convention is that the function has a value for the input argument if the machine halts at the standard output configuration, but if the machine halts in a different configuration or does not halt at all, the function has no value for such argument an it is a partial function (e.g., \cite{boolos-jeffrey-1989}). 

Sustaining that the mind uses algorithms computed by a TM is an a posteriori rationalization but not a causal explanation of mental behavior. Mathematical concepts, notations and metric systems are historical and cultural constructs, that appeared much latter than the machinery used by natural computing, and there is no reason to suppose that the brain and the mind use such constructs, in the same sense that there is no reason to suppose that rationalizations provided by people of the knowledge involved in deploying an ability, such as riding a bicycle, are causal explanations.

Traditional approaches to rationality focus on the general mechanisms and the particular strategies that support rational behavior, and underly the assumption that computing is performed through a general computing device such as the Turing Machine. The present theory suggests to reverse such view and focus on the computing engine that is causal to rational behavior.

%The assumption that people compute standard algorithms is not sound because it is not known whether the brain is a computing engine, and if if were it is not known what are its structural ingredients, what is its standard configuration and which are the interpretation conventions. Indeed, people are very bad to compute algorithms, and when they do, use external aids, such as pencil and paper; computing algorithms is very expensive in terms of memory and time, and cannot be done in real-time by people unless these are very simple. 

Postulating mental algorithms is a very productive metaphor that has contribute to make great progress in Cognitive Science, but interpreting it literally cannot be sustained unless the specification of the mode or modes of natural computing is provided.

The actual natural computing engine is likely to use a highly distributed format where computing is performed by very simple processing units that compute very simple algorithms. The complexity comes from the coordinated computation of such units. The interpretation of such distributed representations as whole units of content is exemplified by minimal algorithms. This view is quite similar to the initial formulation of neural networks \cite{Rumelhart} but its implementation in table computing, for instance, can be done with massive arrays that perform computations in parallel in a few computing steps, and does not need to be reduced to TMs and computed through costly algorithms, as has been the case with artificial neural networks to the present day.

% A further consideration is that the representations on the tape are interpreted by people. The machine manipulate the symbols but it does not know that it is computing a function, neither that the strings on the tape in the standard configuration stand for the argument and value of the function that is being computed. The machine does not know anything. The difference between computing machinery and standard machines or tools is that while the latter are designed and built to produce some useful work, the former does nothing but manipulating representations to be interpreted by people. Computing machinery is a supporting external device, such as paper and pencil, that automatize the symbolic manipulation, and there may be a large kind of devices or modes of computing that provide this support, but the act of computing always ties a mode of computing with an interpretation. This can be stated more categorically: there is no computing without an interpretation. In the case of standard computers the interpretation is performed by people, but in the case of natural computing the agent supporting the mode of computing and the one performing the interpretation is the same. In natural computing interpretation and computing are two sides of the same coin \cite{Pineda-modo-comp}.

\section{Levels of Cognition}
\label{se:rationality}

The cognitive architecture in  Figure \ref{fig:cog-arch} includes the main modules of cognition but some of its components or even full modules can be removed, rendering simpler forms of cognition. Here three main levels are distinguished: 1) the full architecture supporting perception and action, schematic behavior, deliberative thought and memory; 2) the architecture resulting from removing the deliberative thought and the memory modules but preserving the schemes; and 3) the one supporting only basic observations and actions, and rendering reactive behavior.

In a level-2 architecture ``the intentions'' are reduced to the output of the schemes which drives the action directly. This architecture preserves the perceptual and action abilities, and behavior may be refined through training and rehearsing, but lacks the input from deliberative thought and memory, and there is no genuine decision making. Agents that have such an architecture do not have knowledge proper, learning is reduced to training, and behavior is schematic and data-driven.

This kind of agents may be biased due to both the schemes and the training data, and such prejudices could not be modulated because the lack of knowledge. This is a strong limitation in relation to the fully rational architecture where the perceptual and the action abilities are also trained on the basis on empirical data, but the knowledge acquired and learned through language allows for the appearance of values, and the development of an affective logic, and biases and prejudices can be modulated, reduced and even eliminated. 

The level-2 architecture is exemplified by current deep-learning and reinforcement learning models and their applications \cite{Lecun-nature}; for instance, self-driven vehicles such as cars and drones, and even chess, shogi and go playing programs \cite{silver-2018}.

The level-3 architecture only supports basic or ``innate'' observations and motor actions which are connected directly, and agents endowed of these capabilities have very limited or none training capabilities, and deploy mostly reactive behavior. Agents with this kind of architecture do not make interpretations and hence do not communicate.

Agents with architecture level-1 may bypass thought and use perceptual and action abilities directly, which can also be bypassed by reactive behaviors; hence, the level-1 architecture embeds a level-2 and a level-3, and agents with a level-2 architecture embed a level-3 too. The hierarchical embedding, along the lines proposed by Brooks \cite{Brooks}, may be essential for effective interaction with the world.

\section{Principle of Rationality}
\label{sec:rationality}

A computing agent is rational to the degree in which its actions allow it to survive and improve its living conditions, for itself and for other agents in its environment, in the short, middle and long term. A precondition of rationality is that the agent's intentions reflect its own needs and desires; that the decision making engine conforms to the decision making trade-off; that schematic behavior is effective but biases are prevented; that the agent behaves according to the principles of interpretation and action; and that the potential productivity of decisions allows or provides the space for decisions to be enacted. This is referred to here as \emph{The Principle of Rationality}.

Perception produces an interpretation hypothesis about the state of the world; a decision is a hypothesis about the best course of action to achieve a state of the world that the agent believes is needed or desired; the intention underlies such a hypothesis; an action is based on the hypothesis that its consequences will satisfy such intention; and the whole cycle of perception, thought and/or schematic behavior and action is hypothetical. As in evolution --where genetic accidents produce traits that enhance, inhibit or create new capabilities, but only those that provide an advantage are preserved, with the consequent impact on the individual agent and the species-- there is no objective measure or judgement of how rational is behavior but its consequences --benefits or shortcomings-- for the agent and its social and physical environment.

Irrational behavior occurs when decisions and actions harm or diminish the quality of life of the agent and/or the environment. These behaviors may be due to impairments in perception and action, or to limitations in thought and decision making, or to the use of heuristics that result on unfounded actions. 

Rational behavior should conform to the entropy trade-off and the environment should sustain a moderate level of entropy where the potential productivity of decisions is satisfactory or optimal. If the entropy of the computing engine is too low, in relation to the normal level, the agent will act obsessively or according to stereotypes, and will not have the flexibility required to attend the changing demands of the environment; but if the entropy is too high in relation to the normal level, the behavior would not be focused to act productively, as when the attention is impaired. In a sense, a healthy mind and genuine decision making would reflect a moderate or optimal level of entropy.

The present principle of rationality opposes Newell's corresponding principle. In his view, thought is a ``pure process'' performed by symbolic manipulation while in present formulation thought and memory are related to perception and action in a congruent manner, and the agent is placed or grounded in the world.

\section{Brain Entropy}
\label{sec:brain-entropy}

The brain is an entropic machine that sustains a large number of states. A brain state  consists of a reliable pattern of brain activity that involves the activation and/or connectivity of multiple large-scale brain networks, and some states, such as rest, alert and mediation, have been studied \cite{tang-rothbart-posner-2018}. Brain states may have a large number of substates whose ongoing fluctuations influence strongly higher cognitive functions \cite{zagha-2014} and the brain entropy may be related to the number of states that are accessible for brain functioning \cite{saxe-2018}.

The entropy of a brain state can be measured through functional fMRI \cite{wang-2014}. The region under study is divided into voxels, each having a unique value of the blood-oxygen-level-dependent signal (BOLD) at the scanning time, which is correlated with the level of activity of the voxel, and maps of the activity of the brain when people are performing a mental task can be created. The technique consists on registering a sequence or window of  BOLD values over time and computing the indeterminacy of a voxel, which can be characterized through the so-called Sample Entropy or $SampEn$. This methodology has been applied to measure the changes of the brain entropy while a periodic sensorimotor task was performed, with the following results \cite{wang-2014}:

\begin{itemize}
\item Brain entropy provides a physiologically and functionally meaningful brain activity measure.
\item There was an entropy decrease in the visual and sensorimotor brain regions associated to the task in relation to the rest state. 
\item The entropy of the neocortex regions is lower than the entropy of the rest of the brain --cerebellum, brain stem, limbic area, etc.
\item Entropy brain clusters with particular levels of entropy correspond to anatomical or functional areas of the brain.
\end{itemize}

The levels of entropy of the neocortex suggest that the brain conforms to the entropy trade-off. There is a range of entropy values in which interpretation, thought and intentional actions can be effective, as sustained by the free-energy principle and the Bayesian brain hypothesis \cite{Friston-2010}. Other brain structures sustaining higher entropies constitute standard biological machinery that supports but do not make interpretations proper. The cited experiment suggests that the resting state is flexible enough to address the changing demands of the environment, and hence its relatively higher level of entropy, but the entropy lowers when intentional task are performed to achieve the focus and specificity required for performing higher cognitive functions.

The relation between intelligence and brain entropy has been investigated, and a correlation between the entropy at the resting-state and intelligence --measured with verbal and performance IQ tests-- has been found \cite{saxe-2018}: the higher the entropy the higher the IQ. This result is somehow paradoxical because higher IQ is associated to higher leves of indeterminacy, and the experiment is at odds with the entropy decrease associated to intelligence. However, the result can be placed in the perspective of the entropy trade-off which suggests that lower IQ is associated to rigid or predetermined behavior, a higher IQ requieres some degree of indeterminacy, but if the entropy is increased considerably, performance would be decreased accordingly.

Although there is not a well-established and widely accepted notion of brain state, and the current measure of brain entropy may be a gross approximation, the present considerations, in conjunction with the notions of relational-indeterminate computing and the computing entropy, suggest that the neocortex functional regions of the brain are computing entropic engines, and hence they are not Turing Machines, and the brain as a whole is not a Turing Machine. These considerations also suggest that older cortical and subcortical regions should have a very high entropy and may not be considered computing engines.

\section{Technical Challenges and Predictions}
\label{sec:challenges}

The cognitive architecture can be simulated with a TM and in such case the computations are deterministic. If perception, action, schematic behavior, thought and memory, are simulated with standard digital computers, the induction of interpretations, the synthesis of actions, and the decision making process are predetermined. There may be some indeterminacy due to the entropy of the environment, but if the process is carried on by a TM its entropy is zero. However, if the simulation is made with other modes of computing that are intrinsically entropic, such as the relational-indeterminate, or perhaps analogical or quantum computing, or even holography, there may be a level of indeterminacy of the computing agent in addition to the indeterminacy of the environment, and decision making may be free in such degree. 

The present theory poses the challenge of defining and implementing computing processes and associative memories with minimal algorithms that are causal to such kind of behavior.

The present theory suggests as  well a number of hypotheses that may be tested empirically, for instance:

\begin{itemize}

\item Evolutionary psychology and sociology:
\begin{itemize}
\item The potential productivity of decisions: it may be possible to define natural control volumes of human and non-human animal environments, measure their entropies, and count the productive behavioral changes due to communication. If the prediction is sound a $\tau$-profile should be identified. The value of $\tau$ could predict social phenomena, such as the size or degree of organization of a social group.
\item The entropy should be related to the phylogeny of the brain: the oldest structures should have a large level of entropy and the entropy level should decrease according to the more varied and flexible behavior of younger structures; the neocortex, associated to the executive functions, should have the lower levels of entropy in the resting state, as suggested by the results cited above. 
\item The brain entropy of animals in the resting state with more developed neural structures should be lower than the entropy of less developed animals in analogous brain states.
\end{itemize}

\item Neurosciences:
\begin{itemize}
\item The entropy of the brain functional modules: the entropy of brain structures associated to perceptual and motor abilities and schematic behavior should decrease to a lower enough level from the resting state to achieve the determinism associated to concrete interpretations and actions. The brain entropy associated to thought and memory should also decrease from the resting state, but it must remain high enough to allow for creative thinking and memory.
\item Long term memory should have a higher entropy level than working memory in the resting state. Natural forgetting occurs if the entropy exceeds the operational range of long term memory as the stored concepts are confused. If the entropy of working memory gets too low, on the other hand, thinking becomes schematic and behavior is more predictable.
\item Mental disorders: disorders of attentional networks and particular conditions \cite{posner-2007,posner-2019} should be associated to abnormal levels of entropy in relation to the resting state; for instance, obsessive-compulsive disorder should be associated to lower than normal levels; Attention-Deficit/Hyperactivity Disorder (ADHS) to higher than normal levels; and depressive and maniac states should be associated to lower and higher than normal levels respectively.
\end{itemize}

\item Cognitive Psychology:
\begin{itemize}
\item Concrete versus abstract thought: The brain entropy decrease in relation to the resting state should be greater in concrete problem solving than in abstract thinking.
\item Propositional versus Analogical and Diagrammatic Reasoning: The brain entropy decrease of symbolic or propositional reasoning, which is more algorithmic, should be larger than the entropy decrease of diagrammatic or analogical reasoning, that may be more natural.
\item Higher intelligence is associated to a better executive control, hence the entropy level of the central executive should be correlated with intelligence within its operational entropy range. The correlation between higher IQ and higher entropy \cite{saxe-2018} supports this prediction. However, if the entropy of the central executive exceeds its optimal value the IQ should decrease according to the entropy trade-off.
\end{itemize}

\end{itemize}

A more intriguing and fundamental conjecture is that the mind evolved from communicating. Entities that do not communicate are merely reactive and have zero entropy, hence do not make interpretations and may not sense or experience the world. Schematic behavior is the paradigmatic form of experiencing the world somehow unconsciously, that is common in human and non-human animals with a developed enough nervous system; and deliberative thought is a form of experience that involves decision making and anticipating the world, where communication is a more productive behavior, as characterized by the potential productivity of decisions. The decrease of the entropy but taking into account the entropy trade-off may be correlated with the level of experience and consciousness in humans and a great variety of animal species.

\section{Acknowledgments}

The author thanks Gibr\'an Fuentes and Rafael Morales for their help in the design and implementation of the experiment in Section \ref{sec:experiment}. The author also thanks the partial support of grant PAPIIT-UNAM IN112819, M\'exico.

%\nolinenumbers

%\section{Bibliography styles}
%There are various bibliography styles available. You can select the 
%style of your choice in the preamble of this document. These styles are 
%Elsevier styles based on standard styles like Harvard and Vancouver. 
%Please use Bib\TeX\ to generate your bibliography and include DOIs 
%whenever available.
%Here are two sample references: \cite{Feynman1963118,Dirac1953888}.

%\section{References}

\bibliography{references}

\end{document}